\documentclass[review]{elsarticle}
\usepackage{lineno,hyperref}
\usepackage{hyperref}
\usepackage{bbm}
\usepackage{caption}
\usepackage{subcaption}
\modulolinenumbers[5]
\usepackage{float}

\journal{Journal of \LaTeX\ Templates}

\begin{document}
\nolinenumbers

\begin{frontmatter}

\title{Driver Behavior Extraction from Videos in Naturalistic Driving Datasets with 3D ConvNets}
% \tnotetext[mytitlenote]{Fully documented templates are available in the elsarticle package on \href{http://www.ctan.org/tex-archive/macros/latex/contrib/elsarticle}{CTAN}.}

\author{Hanwen Miao, Shengan Zhang, Carol Flannagan}
\address{$\{hwmiao, zshengan, cacf\}$@umich.edu}
\address{University of Michigan, Ann Arbor, USA}


\begin{abstract}
% This template helps you to create a properly formatted \LaTeX\ manuscript.
Naturalistic driving data (NDD) is an important source of information to understand crash causation and human factors and to further develop crash avoidance countermeasures. Videos recorded while driving are often included in such datasets. While there is often a large amount of video data in NDD, only a small portion of them can be annotated by human coders and used for research, which underuses all video data. In this paper, we explored a computer vision method to automatically extract the information we need from videos. More specifically, we developed a 3D ConvNet algorithm to automatically extract cell-phone-related behaviors from videos. The experiments show that our method can extract chunks from videos, most of which (~79\%) contain the automatically labeled cell phone behaviors. In conjunction with human review of the extracted chunks, this approach can find cell-phone-related driver behaviors much more efficiently than simply viewing video.  
\end{abstract}


\end{frontmatter}

% \linenumbers

\section{Introduction}
Research and data on automotive safety have been a high priority for many decades. The Fatality Analysis Reporting System (FARS), a census of all fatal crashes in the U.S., began in 1975 \cite{NHTSA}. This national dataset enabled analysis and tracking of the most severe crashes and began to lead to improvements in vehicle design. The National Automotive Sampling System Crashworthiness Data System (NASS-CDS) was launched in 1979 \cite{sampling_system} and included detailed crash investigations of towaway crashes involving light vehicles. This dataset provided key information about causes of injury and death that enabled research and development into design and mandate of seatbelts, airbags, crush zones, and many other safety systems that have saved tens of thousands of lives.

In more recent years, automotive safety research has advanced understanding of crash causation, which has led to the development of crash avoidance countermeasures. A key source of data to inform this development has been naturalistic driving data (NDD), where participants' vehicles are equipped with a variety of sensors, including cameras (interior and exterior facing), accelerometers, and GPS's, among others. The largest such dataset in the world was collected through the Second Strategic Highway Research Program (SHRP2) Naturalistic Driving Study (NDS) \cite{SHRP2}.  

While NDD have supported hundreds of research papers and advances in crash countermeasures and understanding human factors of drivers, the video data are generally grossly underused. The standard approach to obtaining analyzable information from video data is to use human coders, who watch hours of video and annotate such things as driver activities (including secondary tasks), Safety Critical Events (SCEs) (which are identified using kinematic triggers and verified using video), presence/absence of passengers, use of seat belts or car seats, weather and road conditions, and so on \cite{Precht}\cite{Nipjyoti}\cite{Shan}. Video-coded information is key to understanding driver risk factors (e.g., texting while driving) and pre-crash scenarios and kinematics (through identification of SCEs). This limitation means that, for example, of the 35 million hours of driving in the SHRP2 NDS, the most-used video-based data is around 200 hours that were annotated in the 32,500 20-second clips of baseline driving and around 1500 SCEs (i.e., crashes and near-crashes) \cite{SHRP2}. 

One potential way to dramatically increase the usability of the video data is by using computer vision algorithms to automatically annotate. The field of computer vision has advanced significantly in the last decade. Notably, the simultaneous development of neural network methods and hardware for computational capacity have supported major improvements in algorithm performance. Initially, convolutional neural networks(ConvNet) \cite{Dan} were applied to images and achieved great improvements on some tasks like image classification compared to traditional methods. With its success on images, researchers further proposed 3D convolutional neural networks(3D ConvNet)\cite{C3D} to deal with videos. Inflated 3D ConvNet(I3D)\cite{I3D} represent further developments. Based on ImageNet-pretrained Inception-V1, all the filters and pooling kernels in Inception-V1 are inflated into 3D to add an additional temporal dimension.  These 3D ConvNets were first used for video classification and achieved excellent performance. \cite{chunk-based_classifier} proposed an idea to directly use these 3D ConvNets to temporally localize activities in untrimmed videos.
% Indeed, these algorithms are used extensively for crash avoidance technologies (e.g., Forward Collision Warning) and Automated Driving Systems (ADS) in the field.

To further the usability of NDD with video, such as the SHRP2 NDS, and inspired by \cite{chunk-based_classifier}, this paper explores the possible use of 3D ConvNets to detect cell-phone-related behaviors on the part of drivers. We develop a 3D ConvNet algorithm using NDD from the Integrated Vehicle-Based Safety Systems (IVBSS) Field Operational Test (FOT)\cite{IVBSS}. We then present the performance of that algorithm, discuss some potential use cases, evaluate the use of our algorithm, and identify the performance level needed for each use case.

% \paragraph{Installation} If the document class \emph{elsarticle} is not available on your computer, you can download and install the system package \emph{texlive-publishers} (Linux) or install the \LaTeX\ package \emph{elsarticle} using the package manager of your \TeX\ installation, which is typically \TeX\ Live or Mik\TeX.

% \paragraph{Usage} Once the package is properly installed, you can use the document class \emph{elsarticle} to create a manuscript. Please make sure that your manuscript follows the guidelines in the Guide for Authors of the relevant journal. It is not necessary to typeset your manuscript in exactly the same way as an article, unless you are submitting to a camera-ready copy (CRC) journal.

% \paragraph{Functionality} The Elsevier article class is based on the standard article class and supports almost all of the functionality of that class. In addition, it features commands and options to format the
% \begin{itemize}
% \item document style
% \item baselineskip
% \item front matter
% \item keywords and MSC codes
% \item theorems, definitions and proofs
% \item lables of enumerations
% \item citation style and labeling.
% \end{itemize}

% \section{Related Works}

% The author names and affiliations could be formatted in two ways:
% \begin{enumerate}[(1)]
% \item Group the authors per affiliation.
% \item Use footnotes to indicate the affiliations.
% \end{enumerate}
% See the front matter of this document for examples. You are recommended to conform your choice to the journal you are submitting to.

\section{Video Data}
For this analysis, we used videos from the Integrated Vehicle-Based Safety System(IVBSS) Study \cite{IVBSS}. The IVBSS FOT, conducted in 2009, was targeted at observing drivers' responses to multiple warning systems in an integrated package. There were 108 subjects who each drove one of 8 identical vehicles for 40 days each. The first 12 days consisted of baseline driving during which the systems were turned off.  This was followed by 28 days of driving with the system.  

The IVBSS data are ideal for this study because the entire baseline dataset for all drivers was previously annotated for cell-phone-related behaviors for secondary data analysis purposes(e.g. \cite{Xiong}). Cell-phone events were divided into three categories: cell-phone interaction, dialing a cell-phone, and talking on the cell-phone. The remaining video was labeled "no cell-phone-related behavior." Start and end frames for each cell-phone behavior were identified. 

\begin{figure}[hbt!]
    \centering
    \includegraphics[width=\textwidth]{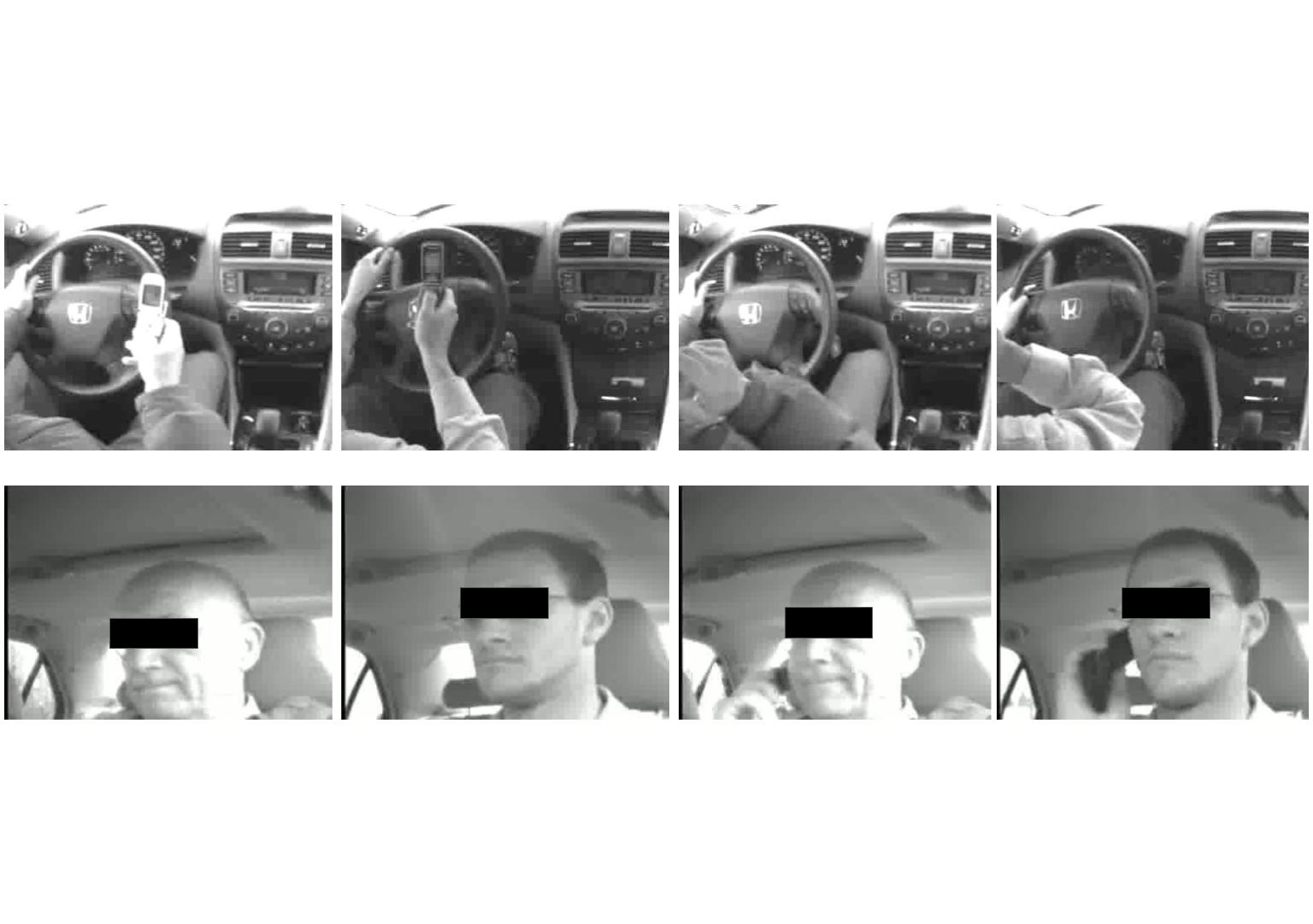}
    \caption{Sample frames from videos containing cell phone behaviors. The first row corresponds to the cabin view and the second row corresponds to the face view. The first two columns correspond to the behavior of interacting with the phone, the last two columns correspond to the behavior of talking on the phone. }
    \label{fig:sample frames}
\end{figure}

There are two camera views in the IVBSS videos: a face camera that centers on the face and the upper shoulders of the driver, and a cabin camera that includes a view of the driver’s hands on or around the steering wheel. The face video frame rate is 10Hz and the cabin video frame rate is 2Hz. Examples of these two views are shown in Figure \ref{fig:sample frames}. It is worth noting that because of the time of data collection for IVBSS, the phones are generally an older "flip" or "candy bar" style rather than a smart-phone style.

In our study, we used the face and cabin video views from 96 drivers throughout the whole baseline period. These drivers all had at least one cell-phone behavior observed and gave permission for their video to be used for viewing and secondary data analysis.

We divided the videos into a training set, a validation set and a test set based on driver ID. Drivers were not reused to prevent the model from benefiting from driver-specific learning. In principle, an algorithm to be used on a specific dataset (e.g., IVBSS or SHRP2) could be allowed to learn the particular drivers in the dataset to improve accuracy on implementation, but for this study, we followed this more challenging protocol. 

The training set was used to train the models; the validation set was used to test the model for the purpose of stopping the training to avoid overfitting, and the test set was used for evaluation of the final models. The proportion of drivers in the training set, validation set and test set were 0.7, 0.2, 0.1 respectively.

\section{Method}
Our method is composed of three modules: clip generation, clip classification and clip aggregation. The clip generation module breaks videos into short overlapped clips. The clip classification module predicts the following items for each clip: the category the clip belongs to and whether the start or the end point of a cell-phone behavior falls into the clip. Finally, clip aggregation recombines consecutive clips labeled with the same behavior to produce whole events of differing lengths with the same behavior label.

\subsection{Clip Generation}
\begin{figure}[hbt!]
    \centering
    \begin{subfigure}[b]{0.5\textwidth}
        \centering
        \includegraphics[width=\textwidth]{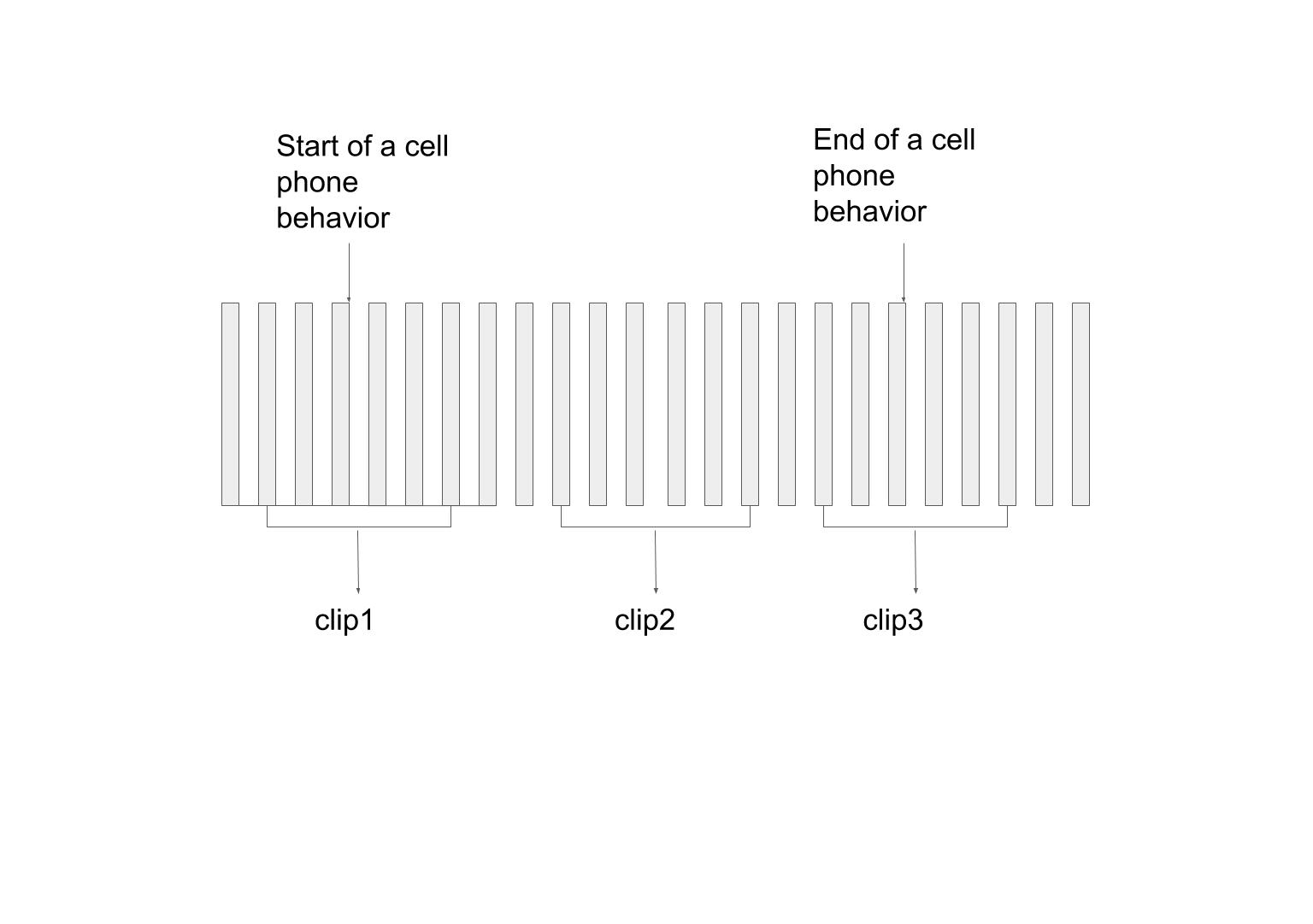}
        \caption{Training stage}
    \end{subfigure} 
    \begin{subfigure}[b]{0.5\textwidth}
        \centering
        \includegraphics[width=\textwidth]{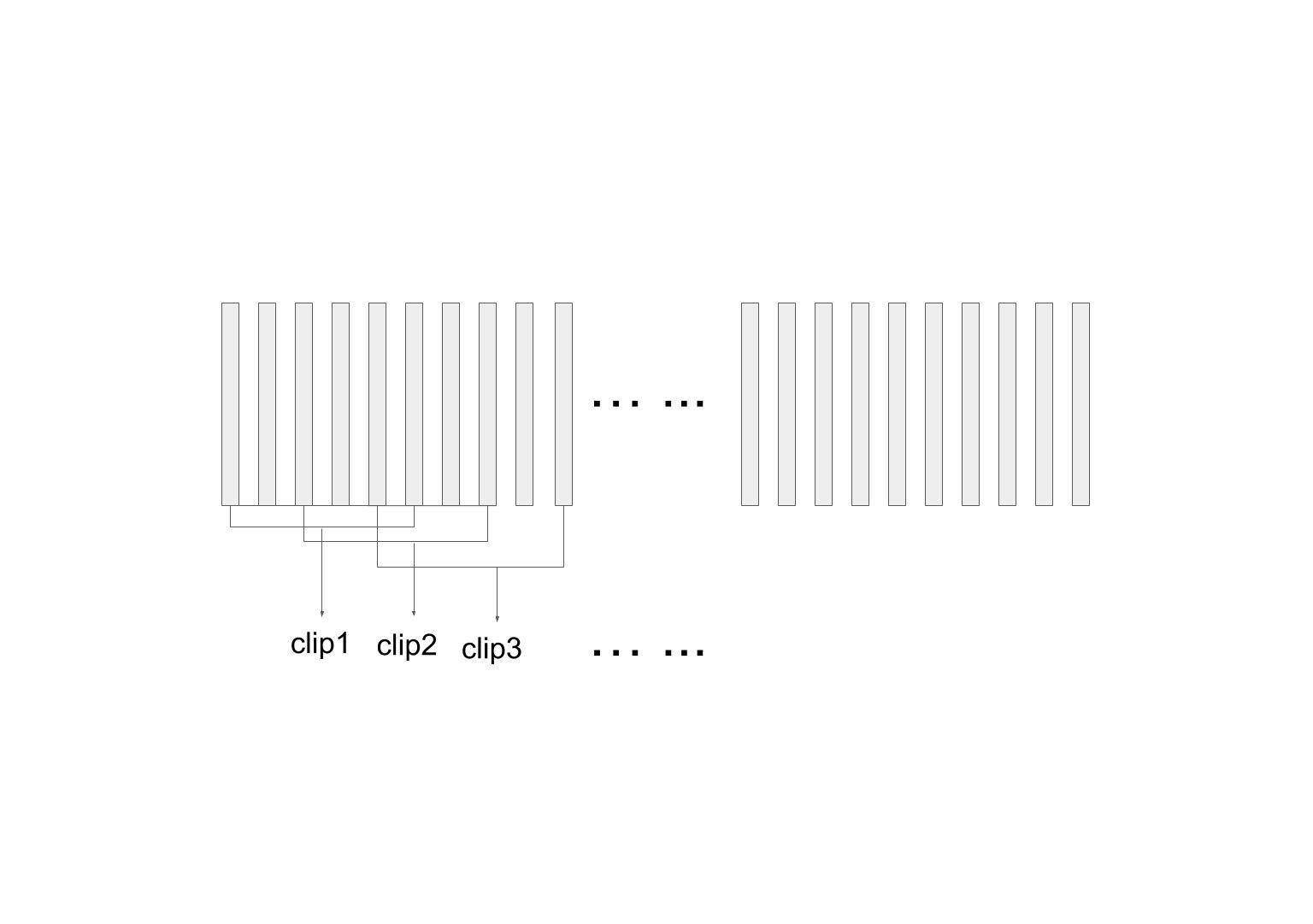}
        \caption{Inference stage}
    \end{subfigure}
    \caption{Clip generation module. In the training stage, we extract three clips for each behavior. In the inference stage, we extract consecutive overlapped clips.}
    \label{fig:clip generation module}
\end{figure}

We first needed to extract short clips from videos. This process was a little different between the training stage and the inference stage. During the training stage, for each video, we extracted three kinds of clips: 1) clips containing the start point of a behavior; 2) clips containing the end point of a behavior; and 3) clips between the start point and the end point. The duration of each clip was 8 seconds. Because of the different frame rates, the clips for cabin views had 16 frames each and the clips for face views had 80 frames each. We downsampled the face clips to 16 frames using interpolation, so that we could stack the features from two feature extractors along the dimension corresponding to the number of filters. For each clip, we provide three labels: class label, start inclusion label and end inclusion label.

During the inference stage, we divided a video into overlapped clips of 8 seconds length. A sliding window of the first 8 seconds was designated as the first clip. Then the window moved by 2 seconds to generate the next clip, and so on. In this way, each 2 seconds of video is part of up to four clips. 

\subsection{Clip Classification}
% \begin{figure}[h]
%     \centering
%     \begin{subfigure}[b]{0.5\textwidth}
%         \centering
%         \includegraphics[width=\textwidth]{figures/Clip Classification Module.jpg}
%         \caption{Clip Classification Module}
%     \end{subfigure} 
%     \begin{subfigure}[b]{0.5\textwidth}
%         \centering
%         \includegraphics[width=\textwidth]{figures/I3D model.jpg}
%         \caption{Inceptive 3D model}
%     \end{subfigure}
%     \caption{Model of the clip classification module. We use the part in the red bounding box as the feature extractor.}
%     \label{fig:clip classification module}
% \end{figure}
\begin{figure}[hbt!]
    \centering
    \includegraphics[width=\textwidth]{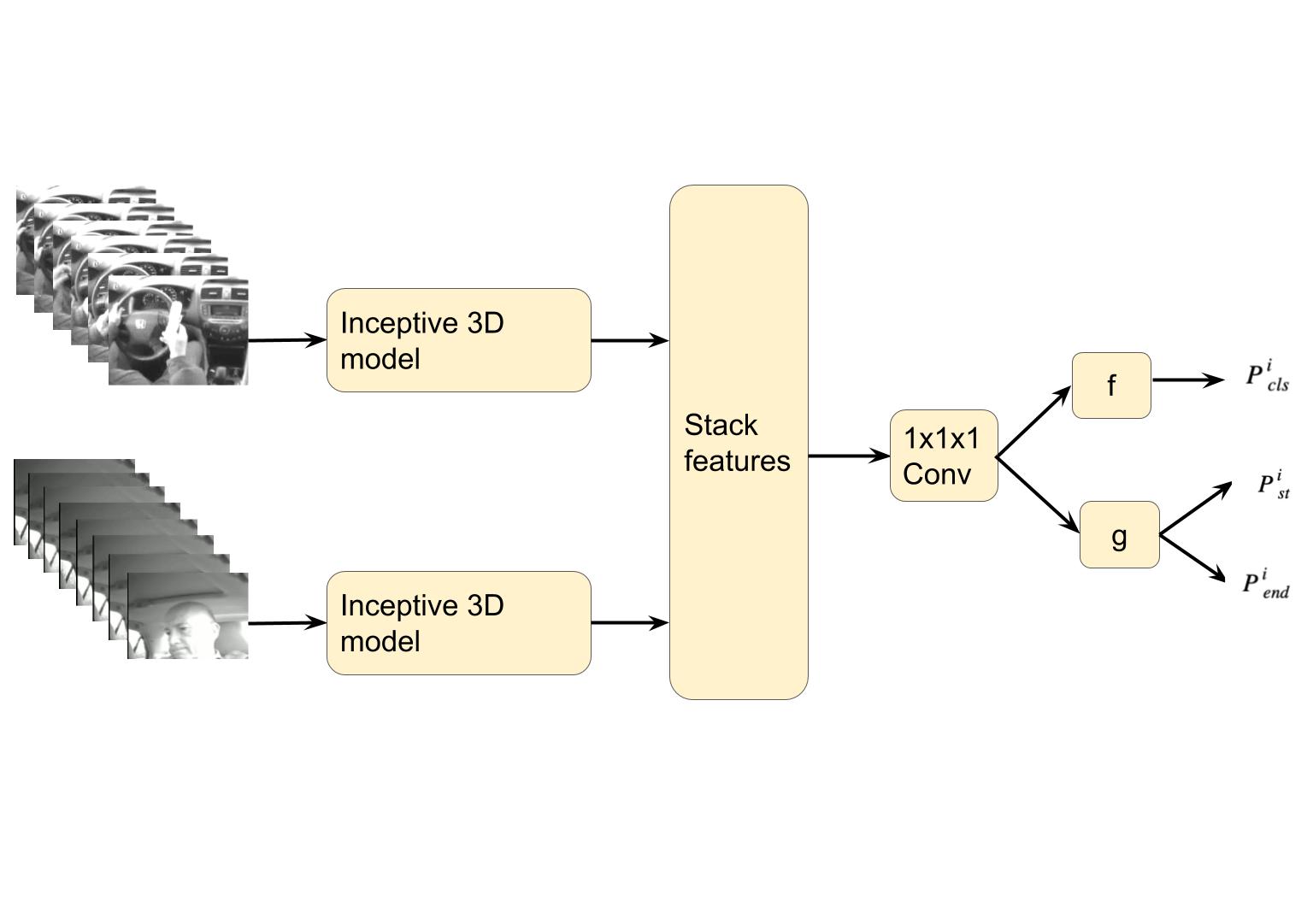}
    \caption{Clip Classification Module}
    \label{fig:clip classification module}
\end{figure}

The clip classification module was built based on an existing video classification model: Two-stream Inflated 3D ConvNet(I3D). The structure of the module is shown in Figure \ref{fig:clip classification module}. The detailed structure of I3D can be seen in Figure 3 in \cite{I3D}. In the original paper, the I3D model receives RGB inputs or flow inputs and predicts the activity class. We used all of the layers except the last "1$\times$1$\times$1 Conv" layer as a feature extractor. Images from both cabin view and face view were fed into two separate feature extractors. Then we stack the extracted features and fed them into a new classifier layer. For this classifier layer, in addition to the output branch for the behavior classification task, we add two branches for the clip inclusion task, which predicts whether the start or the end point of a cell-phone-related behavior falls into the clip.

We trained the behavior classification task with cross entropy loss:
\begin{equation}
    \mathcal{L}_{cls}^{i} = -log(\frac{exp(p_{cls}^{i}[class_{i}])}{\sum_{j}^{N} exp(p_{cls}^{i}[j])})
\end{equation}
where $\mathcal{L}_{cls}^{i}$ is the behavior classification loss of the $i$th clip, $p_{cls}^{i}$ is a vector of class scores, $class_{i}$ is the actual class the $i$th clip belonging to and N is the total number of classes.

We trained the clip inclusion task with binary cross entropy loss:
\begin{equation}
    bce(x, y) = -(ylogx + (1-y)log(1-x))
\end{equation}
\begin{equation}
    \mathcal{L}_{inc}^{i} = \frac{1}{2}(bce(p_{st}^{i}, y_{st}^{i})+bce(p_{end}^{i}, y_{end}^{i}))
\end{equation}
where $\mathcal{L}_{inc}^{i}$ is the inclusion loss of the $i$th clip, $p_{st}^{i}$ and $p_{end}^{i}$ correspond to the predicted scores of the inclusion of start point and end point in the $i$th clip, $y_{st}^{i}$ is set to 1 if the start point falls into the $i$th clip and is set to 0 otherwise. Same to $y_{end}^{i}$.

So our final loss can be written as:
\begin{equation}
    \mathcal{L}_{total} = \mathcal{L}_{cls} + \lambda \mathbbm{1}_{a\geq1} \mathcal{L}_{inc}
\end{equation}
where $\mathbbm{1}_{a\geq1}$ is an indicator function which is equal to 1 when the ground truth action class label is greater than 1, and is equal to 0 otherwise. The class label for 'no cell-phone-related behavior' is always set to 0. $\lambda$ is a parameter controlling the weight of class inclusion loss in the total loss and is set to 0.2 in our experiment.

\subsection{Clip Aggregation}
After the clips were classified, we used the outputs for clip aggregation. 
\subsubsection{Rough Aggregation}
The simplest way to aggregate is to merge adjacent clips of the same class. Because clips are overlapped, to get better aggregation result, we first predicted the class of each 2-second clip via majority vote. Each 2-second clip was included in at most four adjacent 8-second clips in our method, and we used the modal class as the final labeled class. This produces a single class for each 2-second clip.  If $X_{i}$ and $X_{j}$ are two adjacent 2-second clips of the same class, we then merged them into a longer clip. In addition, if $X_{i}$ and $X_{j}$ had the same label, we merged them as long as they were no more than 4 seconds apart, even if the intervening clips had different labels. The following describe specific steps:
\begin{itemize}
    \item Step1: aggregate adjacent clips with the same label into a longer clip and save the longer clips with the same label into a list.
    \item Step2: for each list, if there are no more than 4 seconds apart between two adjacent longer clips, they are further merged.
    \item Step3: for all clips, if a shorter clip is covered by a longer clip, only the longer clip is kept. If two clips have a little overlapped part at the boundary, to avoid the conflict, the overlapped part is discarded.
\end{itemize}

\subsubsection{Refined Aggregation}
Simply merging all neighboring clips can achieve satisfactory performance, but we can also use the predicted scores for inclusion (i.e., start and end points) from the clip classification module to achieve better performance. First, we filtered out the clips whose predicted scores for the inclusion branch are higher than their neighbors. We call these peaks. Then if there were no more than 4 seconds between two peaks, we kept only the higher peak. After these two steps, we set a threshold.  If the peak was higher than the threshold $\Theta$, we kept the peak. After these three steps, the kept peaks are the clips with a high probability of containing the start or the end of a behavior.

We used these peaks to refine the rough aggregation. In Step 2 of rough aggregation, if there were no more than 4 seconds between two adjacent longer clips and there were no peaks between them, they were merged. In this case, we avoided merging a wrongly classified short clip to a longer clip, and we also avoided having a small overlapped part at the boundary.

\subsection{Training}
We trained the clip classification module with PyTorch. We first fine-tuned the pretrained I3D model on our videos with only the cabin view and class classification branch. Then we trained the whole model using stochastic gradient descent(SGD) with learning rate 1e-2, momentum 0.9, and weight decay 1e-5. We trained the model for a total of 30 epochs, lowering the learning rate by 0.1 after every 10 epochs. We used a batch size of 8.

\section{Experiments \& Evaluation}
\subsection{Dataset}
At first, we extracted 6934 video clips from the videos to generate the training dataset, which included 1044 'dialing the phone', 2853 'interacting with the phone', 2399 'talking on the phone,' and 638 'no cell-phone related behavior'. Later we built a more evenly distributed training dataset of 6933 clips, which includes 2311 'interacting with the phone', 2311 'talking on the phone' and 2311 'no behavior'. 

% Later we built a more evenly distributed training dataset of 9282 clips, which includes 3094 'interacting with the phone', 3094 'talking on the phone' and 3094 'no behavior'.
% We built the validation dataset and test dataset in the similar way. There are 2069 clips in the validation dataset and 1413 clips in the test dataset.

\subsection{Experiments}
We conducted several experiments and compared their performance. We first trained our model with 4 classes and 2 views. We observed the confusion matrix and found that it was difficult for the model to distinguish 'dialing the phone' from 'interacting with the phone'. We then merged 'dialing the phone' with 'interacting with the phone'. We also trained the model with each of the individual views to identify the contrbutions of each. Finally, the dataset was unequally distributed, so we created a new same-sized training dataset in which each of the three classes had an equal number of clips. All of these models were tested on the same (unequally distributed) test dataset.

\subsection{Quantitative Analysis}
\subsubsection{Accuracy for Class Classification Task}
We first evaluated our model with the class classification accuracy on those clips in the validation and test dataset. As shown in Table \ref{table1}, using both cabin view and face view can achieve better classification accuracy comparing to using just one view, which also tells us that the model utilizes information from both views to detect the cell-phone behaviors.  We can also see that classifying clips with 3 classes can achieve better classification accuracy comparing to classifying clips with 4 classes. Finally, the equally distributed training dataset produced nearly equivalent performance on the test data as the unequally-distributed training dataset, in spite of the test data being also unequally distributed.
\begin{table}[hbt!]
    \begin{center}
    \begin{tabular}{|c|c|c||c|c|}
    \hline
    Camera Views & Classes & Training Dataset & Validation Accuracy & Test Accuracy\\
    \hline
    Two Views & 3 Classes & Unevenly Distributed & 0.788  & 0.794 \\
    Two Views & 4 Classes & Unevenly Distributed & 0.627  & 0.637 \\
    Cabin View & 3 Classes & Unevenly Distributed & 0.736  & 0.721\\
    Face View & 3 Classes & Unevenly Distributed & 0.715 & 0.689 \\
    \hline
    Two Views & 3 Classes & Evenly Distributed & 0.759  &  0.790  \\
    \hline
    \end{tabular}
    \end{center}
    \caption{Comparison of the performance of the model with different numbers of views, different classes and different training sets.}
    \label{table1}
\end{table}

\subsubsection{Confusion Matrix for Class Classification Task}
\begin{figure}[hbt!]
    \centering
    \begin{subfigure}[b]{0.4\textwidth}
        \centering
        \includegraphics[width=\textwidth]{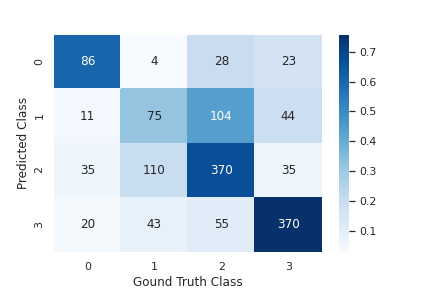}
        \caption{2 views,4 classes, unevenly distributed}
    \end{subfigure}
    \begin{subfigure}[b]{0.4\textwidth}
        \centering
        \includegraphics[width=\textwidth]{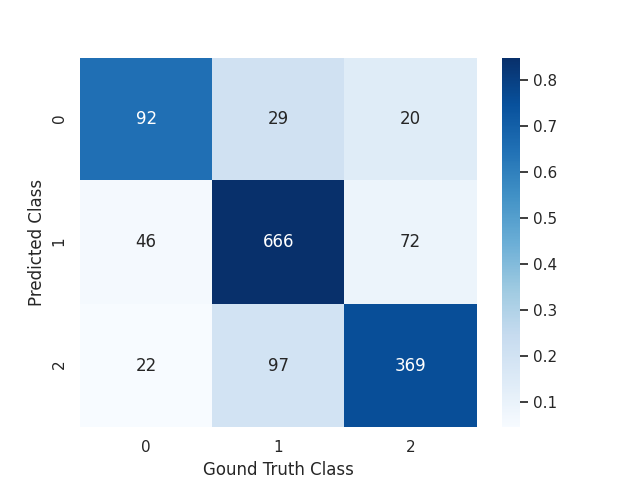}
        \caption{2 views,3 classes, unevenly distributed}
    \end{subfigure} 
    \begin{subfigure}[b]{0.4\textwidth}
        \centering
        \includegraphics[width=\textwidth]{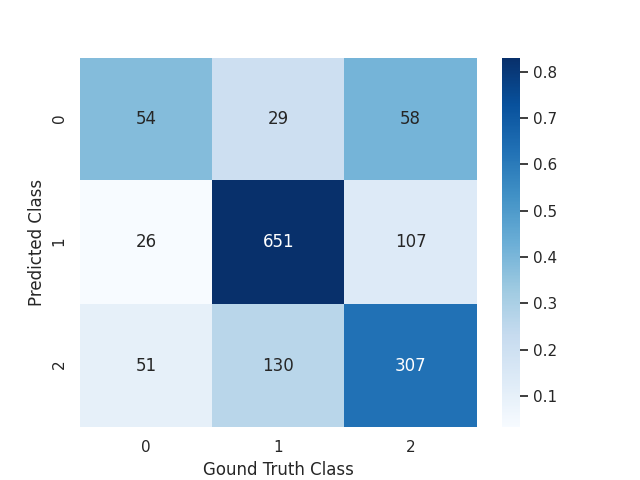}
        \caption{cabin view,3 classes, unevenly distributed}
    \end{subfigure}
    \begin{subfigure}[b]{0.4\textwidth}
        \centering
        \includegraphics[width=\textwidth]{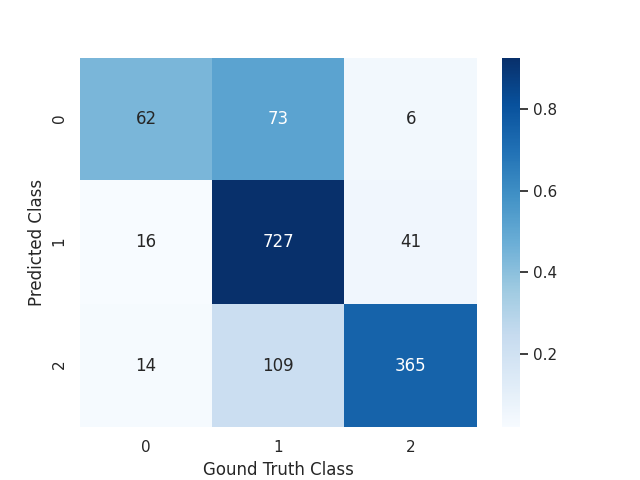}
        \caption{face view,3 classes, unevenly distributed}
    \end{subfigure}
    \begin{subfigure}[b]{0.4\textwidth}
        \centering
        \includegraphics[width=\textwidth]{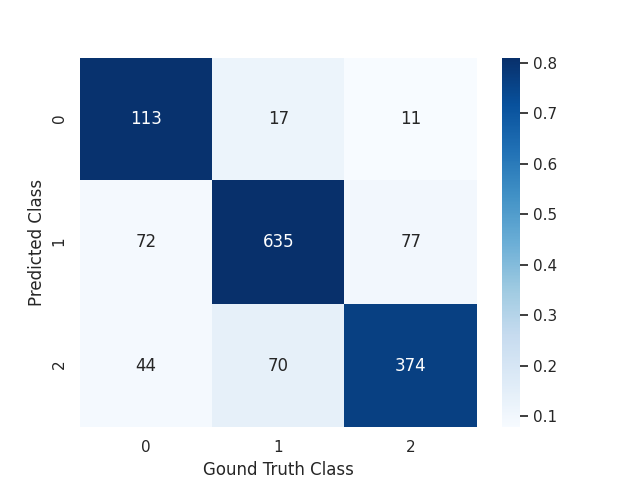}
        \caption{2 view,3 classes, evenly distributed}
    \end{subfigure}
    \caption{Confusion Matrices on the same test dataset. Three classes are 'no activity','interacting/dialing phone','talking on the phone'. Four classes are 'no activity','dialing the phone','interacting with the phone', 'talking on the phone'. The color of each cell corresponds to the row percent.}
    \label{fig:confusion matrices}
\end{figure}

We used confusion matrices to further visualize the performance of classification. From Figure \ref{fig:confusion matrices}(a), it is evident that the model did not distinguish the behavior of dialing the phone from interacting with the phone well. This led us to collapse dialing and interacting into one "interacting/dialing" category. From Figure \ref{fig:confusion matrices}(c), we observe that the model often misclassifies 'no activity' into 'talking on the phone' with only cabin view. Comparing Figure \ref{fig:confusion matrices}(b) and Figure \ref{fig:confusion matrices}(d), we can see that if we train the model on an unevenly distributed dataset, the model tends to misclassify the clip into 'interacting with the phone' when it misclassifies a clip. That is, it learns to select the most common category more readily and thus produces asymmetrical confusion matrices.

\subsubsection{ROC Curves and Precision-Recall Curves for Inclusion Prediction Task}
\begin{figure}[hbt!]
    \centering
    \begin{subfigure}[b]{0.4\textwidth}
        \centering
        \includegraphics[width=\textwidth]{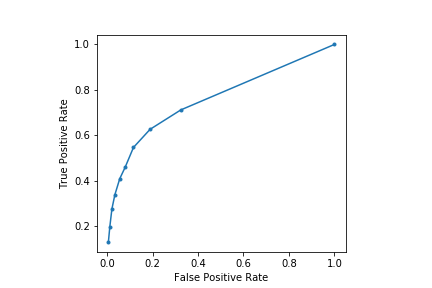}
        \caption{ROC curve for start inclusion prediction}
    \end{subfigure}
    \begin{subfigure}[b]{0.4\textwidth}
        \centering
        \includegraphics[width=\textwidth]{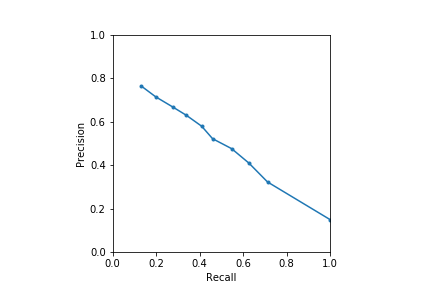}
        \caption{Precision-Recall curve for start inclusion prediction}
    \end{subfigure}
    \begin{subfigure}[b]{0.4\textwidth}
        \centering
        \includegraphics[width=\textwidth]{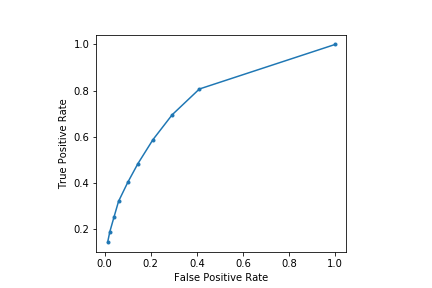}
        \caption{ROC curve for end inclusion prediction}
    \end{subfigure}
    \begin{subfigure}[b]{0.4\textwidth}
        \centering
        \includegraphics[width=\textwidth]{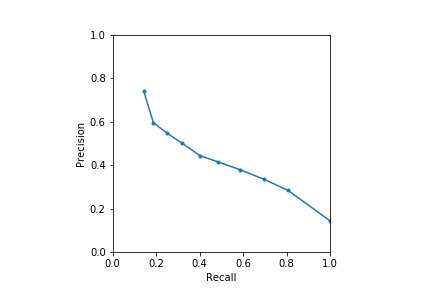}
        \caption{Precision-Recall curve for end inclusion prediction}
    \end{subfigure}
    \caption{ROC curves and Precision-Recall curves for start and end inclusion prediction. Points were calculated based on the thresholds from 0 to 0.9 with a stepsize of 0.1. }
    \label{fig:ROC_curves}
\end{figure}

To evaluate the inclusion branch, we selected 77 videos from the validation and test datasets. Each video contained more than two cell phone behaviors. For each video, we extracted 8-sec clips with 6 seconds overlapped between two adjacent clips and predicted the inclusion scores for each clip. Finally, we tested different thresholds from 0 to 0.9 to produce Receiver Operating Characteristic (ROC) and Precision-Recall (P-R) Curves, which are shown in Figure \ref{fig:ROC_curves}. The ROC curves indicate that the algorithm is generally weak, but better than chance, at detecting start and end points. The P-R curves, especially for start points are linear, indicating that there is no optimal threshold.  

\subsubsection{Temporal Prediction Accuracy}
The final goal of our method is to predict cell-phone behaviors temporally in a continuous video. To evaluate our method, we selected 77 videos from the validation and test datasets. Each video contained more than two cell phone behaviors. For each video, we produced several predicted chunks aggregated from short clips. For each chunk, we calculated temporal prediction accuracy as:
\begin{equation}
    ACC(C_{i}) = \frac{\sum_{j}^{N}L(intersection(GT_{j},C_{i}))}{L(C_{i})}
\end{equation}
where $C_{i}$ is a predicted chunk, $GT_{j}$ is a ground truth chunk of the same class as $C_{i}$, $intersection(GT_{j},C_{i}$ is the overlapped part between two chunks, N is total number of ground truth chunks of the same class. This accuracy measure indicates what percentage of each chunk of video labeled, for example, "talking," is, on average, talking.

As shown in Table \ref{table3}, with rough aggregation, an average of 69\% of predicted chunks of interacting with the phone are actually such behavior and 85\% of predicted chunks of talking on the phone are actually such behavior. The statistics using refined aggregation are 71\% and 86\%, respectively.  The use of start- and endpoint prediction improves the temporal prediction accuracy slightly.

\begin{table}[h]
    \begin{center}
    \begin{tabular}{|c|c||c|c|}
    \hline
    Aggregation Type & Threshold & Interacting with the phone & Talking on the phone\\
    \hline
    Rough Aggregation &  & 0.688  & 0.846 \\
    Refined Aggregation & 0.8 & 0.706  & 0.861 \\
    % Refined Aggregation & 0.7 & 0.714  & 0.862 \\
    % Refined Aggregation & 0.6 & 0.717  & 0.863 \\
    % Refined Aggregation & 0.5 & 0.719  & 0.866 \\
    \hline
    \end{tabular}
    \end{center}
    \caption{Average temporal prediction accuracy of each behavior. Threshold are used to select peaks.}
    \label{table3}
\end{table}

\subsection{Quantitative Analysis}
\subsubsection{Inclusion Prediction}
\begin{figure}[hbt!]
    \centering
    \begin{subfigure}[b]{0.8\textwidth}
        \centering
        \includegraphics[width=\textwidth]{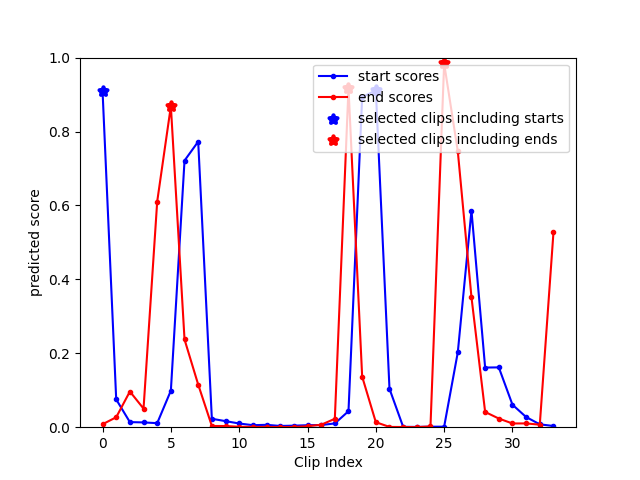}
    \end{subfigure}
  %  \begin{subfigure}[b]{0.8\textwidth}
  %      \centering
  %      \includegraphics[width=\textwidth]{figures/plot_Cabin_078_0007_16870.png}
  %  \end{subfigure}
    \caption{Plots of predicted scores from the inclusion branch, stars are the selected peaks with the threshold of 0.8}
    \label{fig:clip inclusion scores}
\end{figure}

Figure \ref{fig:clip inclusion scores} shows the inclusion branch scores for a series of clips as an example. Blue shows start scores, and red shows end scores. Stars indicate scores that exceeded a threshold of 0.8 for predicting the presence of a start or end. 

\subsubsection{Final Aggregation}
\begin{figure}[hbt!]
    \centering
    \begin{subfigure}[b]{\textwidth}
        \centering
        \includegraphics[width=\textwidth]{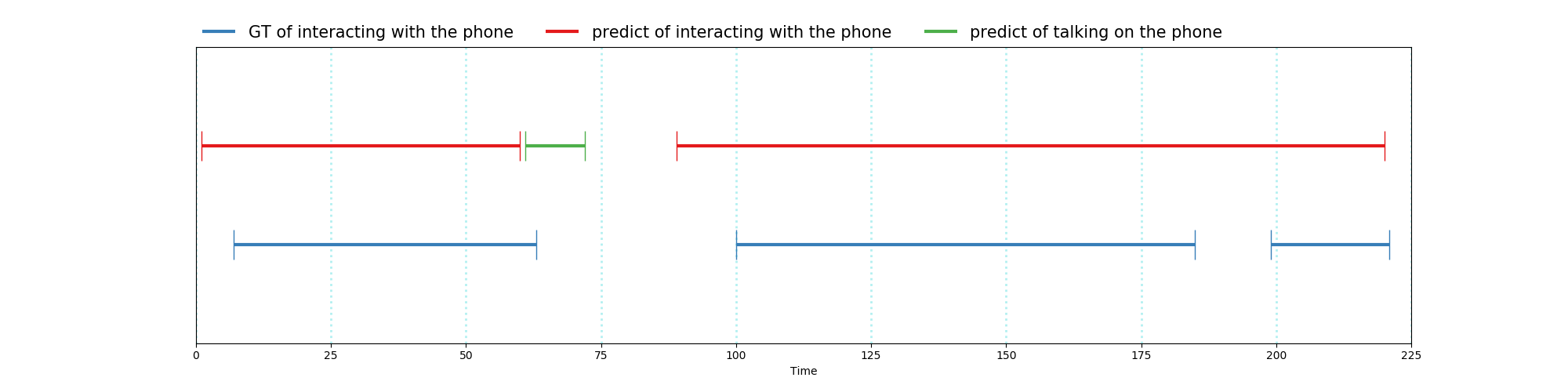}
        \caption{result from rough aggregation}
    \end{subfigure} 
    \begin{subfigure}[b]{\textwidth}
        \centering
        \includegraphics[width=\textwidth]{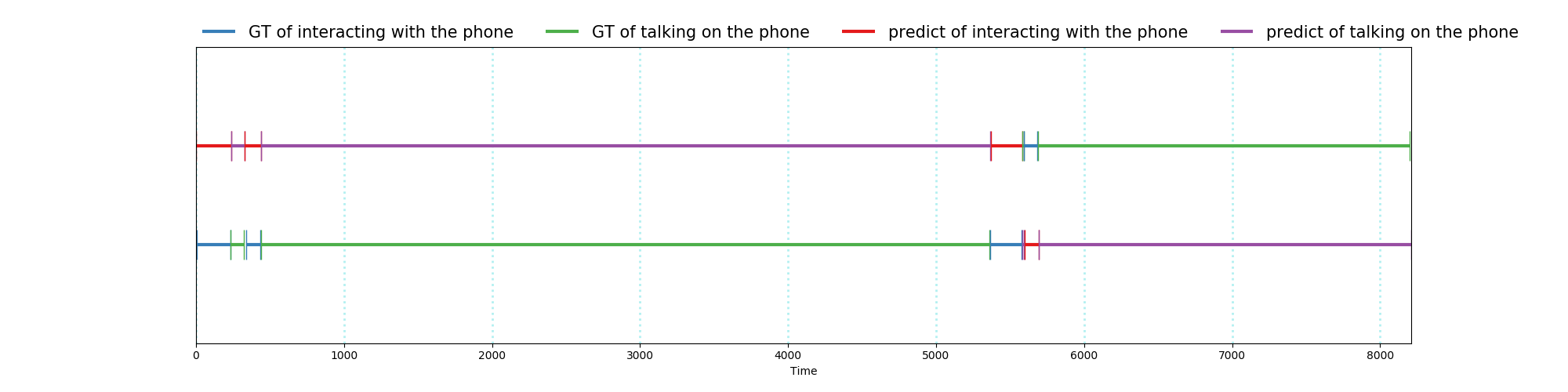}
        \caption{result from rough aggregation}
    \end{subfigure}
    \begin{subfigure}[b]{\textwidth}
        \centering
        \includegraphics[width=\textwidth]{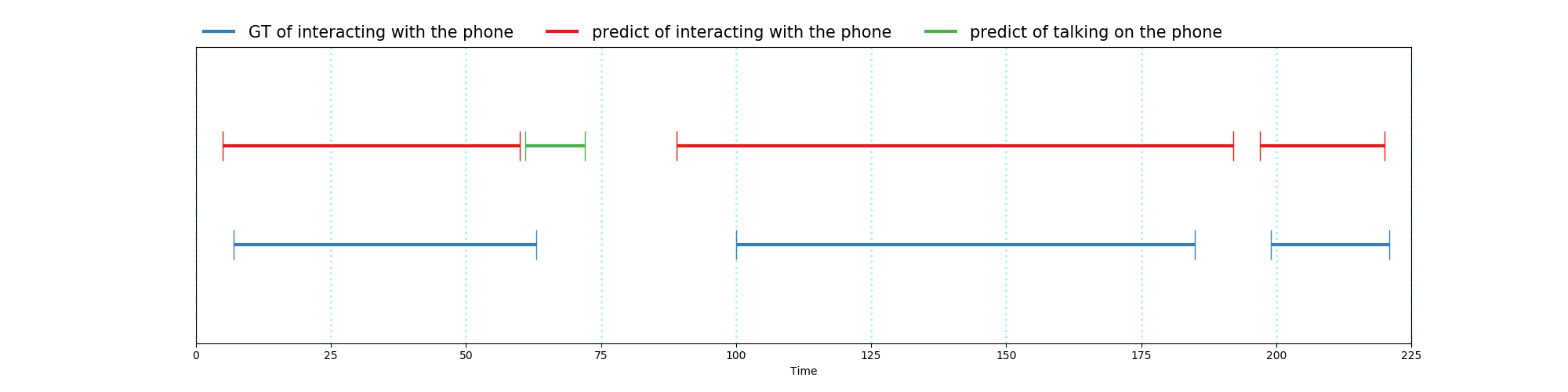}
        \caption{result from refined aggregation}
    \end{subfigure}
    \caption{Visualization of final results from clip aggregation module. (a) and (c) are the same video.}
    \label{fig:temporal localization}
\end{figure}

Figure \ref{fig:temporal localization} shows some final results of our methods. From (a) and (b), we can see that rough aggregation with only class label can perform well. Comparing (a) and (c), we can see that refined aggregation avoids to merge two clips which don't belong to the same behavior.

\section{Discussion \& Future Work}
This work explores the development and use of a 3D ConvNet to identify instances of cell-phone interactions by drivers in the IVBSS dataset. The baseline-driving portion of the dataset had been previously labeled for three cell-phone-related behaviors (dialing, interacting, and talking) based on two camera views (cabin and face). Using these labels, we segmented the videos into 8-second clips and designed a series of algorithms to identify the behavior(including "no activity," which represents none of the above) and to identify the time of the start and end of the activity.

The initial algorithm used the events in proportion to their occurrence in the dataset and used all four behavior categories. This algorithm did not distinguish well between the interacting and dialing categories. We noted that these two behaviors were labeled based on whether they were followed by the behavior of talking on the phone immediately. Because our model is just provided with short clips, it loses the context information. This might be addressed in a future effort that would include some additional context (e.g., in the form of activity order and a secondary algorithm that distinguished between dialing and interacting).

To address this issue, we combined the two confusible categories into a single interacting/dialing category. The resulting algorithm using uneven category membership tended to favor the most common category (interacting/dialing). To address this issue, we reselected clips so that they had the same total number but were evenly distributed among the three categories. This final approach produced an algorithm with the same overall accuracy of 79\% on the test as the original 3-category algorithm but an even confusion matrix that did not favor one category over another. We used this algorithm for further explorations.

While the algorithm does a reasonable job at identifying the activity for 8-second clips, we were interested in identifying longer instances of cell-phone-related activities. This was achieved by aggregating clips based on the predicted class label and the predicted start and end scores. The inclusion models required a threshold to be set, and based on the linear precision-recall curve, there is no optimal boundary. We chose 0.80 to create longer clips. However, the inclusion task performance was somewhat disappointing on its own. In our case, behaviors are always close to each other, so it may be a hard task to distinguish the boundary.

Using our clip aggregation approaches, we find that rough aggregation with only class labels can achieve satisfactory performance and refined aggregation has marginally better performance. If we can get better results from inclusion task, the final result could be much better.

On the whole, our algorithm could reasonably be used in the context of searching for cell-phone use events. This is in contrast with allowing the algorithm to label clips, including start and endpoints, with respect to the three behaviors (including no activity). In a sense, this use case, where manual review is not required, would be the gold standard use case, maximizing benefit. However, an accuracy of 79\% means that many clips would be mislabeled and would add error to analyses that use the labels without review. 

Nonetheless, the algorithm would substantially increase the efficiency of review in search of certain cell-phone-related behaviors. According to NHTSA, 0.6\% of drivers were visibly manipulating a cell phone and 5.0\% were using a handheld cell phone in a 2009 observational study \cite{NHTSA}\cite{traffic}. This estimate should approximate the rates (in time) at which IVBSS drivers engage in cell-phone-related activities. SHRP2, which was conducted 2 years later, should be similar. If 79\% of the clips selected by our algorithm contain cell-phone related behaviors, then it would be substantially more efficient to review those than to review all video, where about 6\% of clips are expected to contain behaviors of interest. This means that approximately 13 times (0.79/0.06) more cell-phone activity clips can be found in the same amount of review time using the 3DConvNet approach. Even given the time to train the model, this would be an efficient approach to gathering instances for certain research questions in which a larger sample of events where drivers engage in specific activities is needed.

% Here are two sample references: \cite{Feynman1963118,Dirac1953888}.

% \section*{References}
\newpage
\bibliography{mybibfile}

\end{document}